\documentclass[1p,11pt,times]{elsarticle}

\usepackage{ecrc}
\usepackage{epsfig}

\usepackage[cmex10]{amsmath}
\usepackage{amssymb}
\usepackage{amsthm}
\usepackage{algorithmic}
\usepackage{array}     	
\usepackage{subfigure}
\usepackage[normalem]{ulem}  
\usepackage[linesnumbered, ruled, vlined]{algorithm2e}
\usepackage{verbatim}  	
\usepackage{tabularx}  	
\usepackage{multirow}  	
\usepackage{booktabs} 	
\usepackage[table]{xcolor}  
\usepackage[bookmarks]{hyperref}
\usepackage{url}
\usepackage{epstopdf}

\jid{procs}

\usepackage{amssymb}

\usepackage[figuresright]{rotating}

\begin{document}

\begin{frontmatter}




\title{\bf Skeleton Matching based approach for Text Localization in Scene Images}


\author[rvt]{B.H.Shekar\corref{cor1}}
\author[els]{Smitha M.L.}

\cortext[cor1]{Corresponding author}                             
                                                            
\address[rvt]{Department of Computer Science, Mangalore University, Mangalore, Karnataka, India. \\ Email:
bhshekar@gmail.com}
\address[els]{Department of Master of Computer Applications, KVG College of Engineering, Sullia, Karnataka, India.\\
smithaml.urubail@gmail.com}                                         
\begin{abstract}
In this paper, we propose a skeleton matching based approach which aids in text localization in scene images. The input image is preprocessed and segmented into blocks using connected component analysis. We obtain the skeleton of the segmented block using morphology based approach. The skeletonized images are compared with the trained templates in the database to categorize into text and non-text blocks. Further, the newly designed geometrical rules and morphological operations are employed on the detected text blocks for scene text localization.  The experimental results obtained on publicly available standard datasets  illustrate that the proposed method can detect and localize the texts of various sizes, fonts and colors.
\end{abstract}

\end{frontmatter}

\section{Introduction}
\label{section:Introduction}
Text localization in document/scene images and video frames aims at designing an advanced optical character recognition (OCR) systems. However, the large variations in text fonts, colors, styles, and sizes, as well as the low contrast between the text and the complicated background, often make text detection extremely challenging. The researcher's experimental results on such complex text images/video reveals that the applications of conventional OCR technology leads to poor recognition rates. Therefore, efficient detection and segmentation of text blocks from the background is necessary to fill the gap between image/video documents and the input of a standard OCR system.  	

The text-based search technology acts as a key component in the development of advanced image/video annotation and retrieval systems. In general, the methods for detecting text can be broadly categorized into five classes based on the features associated with the text. The various approaches are connected component analysis(CCA) approach, edge-based, corner-based, texture-based and stroke-based approaches. The connected component analysis approaches \cite{zhong2000automatic},\cite{Zhan2006} assume that the pixel in text regions have homogeneous color, intensity, texture. The color-based methods are simple and are suitable only to simple background. The edge-based approaches\cite{Lienhart2002},\cite{Jain1998} require text to have a reasonably high contrast to the background in order to detect the edges. These methods often encounter problems with complex backgrounds and produce many false positives. The corner-based methods\cite{Li200},\cite{Ye2004} extract corner features to detect the text in images. The corner based methods are more effective but detecting the corners generally is a time consuming task. The texture-based methods\cite{Liu2005},\cite{hua2004automatic} assume text regions to have some kind of special textures. The texture-based methods are time-consuming and sometimes are influenced by the fonts and styles of characters. The stroke-based method\cite{epshtein2010} captures the intrinsic characteristics of text strokes so that the better detection results have been obtained even in complex background.

Although the text recognition in documents is satisfactorily addressed by state-of-the-art OCR systems, the text localization and recognition in images of real-world scenes has received significant attention in the last decade. Hence, the scene text localization and recognition is still an open problem. In this context, we propose a new approach for text localization  in scene images. Our method aims to detect the text in the input image by performing certain preprocessing. Further, the preprocessed image undergoes segmentation, skeletonizing, template matching, classification and localization. The remaining part of the paper is organized as follows. The proposed approach is discussed in section 2. Experimental results and comparison with other approaches are presented in section 3 and conclusion is given in section 4.
\section{Proposed Methodology}
\label{section:algo1}
\begin{figure}[hbtp]
\centering
\includegraphics[scale=0.575]{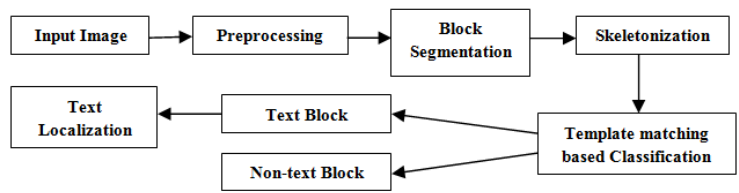}
\caption{Flowchart of the proposed system}
\label{fig1}
\end{figure}

The flowchart of the proposed text detection and localization approach is shown in Fig.~\ref{fig1}. The details of each processing blocks are discussed below. 

\subsection{Preprocessing \& Segmentation} 
\begin{figure}[hbtp]
\centering
\includegraphics[scale=0.375]{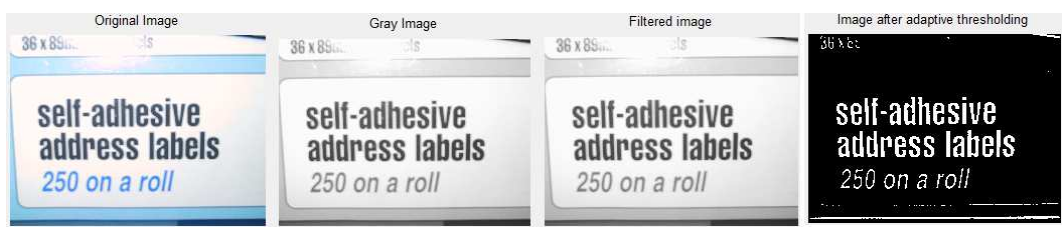}
\caption{Results of Binarization}
\label{fig2}
\end{figure}

In this phase, the input image is segmented into small blocks using connected component analysis. The given input image is first converted into a gray image and median filtering is employed to remove the noise in the resultant image. We employ an efficient binarization approach over the filtered image to get a binarized image. If the background in the image is relatively uniform then a global threshold value is used to binarize the image by pixel-intensity. If there is large variation in the background intensity then adaptive thresholding (local or dynamic thresholding) may produce better results. The conventional thresholding operator uses a global threshold for all pixels whereas adaptive thresholding changes the threshold dynamically over the image. Adaptive thresholding is a form of thresholding that takes into account spatial variations in illumination. The filtered image is further binarized using local adaptive thresholding which selects an individual threshold for each pixel based on the range of intensity values in its local neighborhood  as shown in Fig.~\ref{fig2}. 
 \\For segmentation purpose, we perform connected component analysis, a technique that scans and labels the pixels of a binarized image into  components based on pixel connectivity. The connected component labeling detects the connected regions in the binary images. The binarized image is now segmented into blocks using connected component analysis.   
\subsection{Skeletonization}
Skeletonization\cite{zhong2000automatic},\cite{Zhan2006} is a morphological operation that is used to remove selected foreground pixels from binary images. Skeletonization is normally applied only to a binary image and it produces another binary image as its output. The skeleton of the character image is important for its detection. Hence, the redundant information may be rejected using skeletonization\cite{You2007}. We now find the skeleton of the segmented block which leads to an image with single pixel width maintaining the basic shape of the image. The resultant images are resized to form templates of fixed size. These unknown templates are used  to compare with the known templates in the template database. 
\begin{figure}[hbtp]
\centering
\includegraphics[scale=0.3]{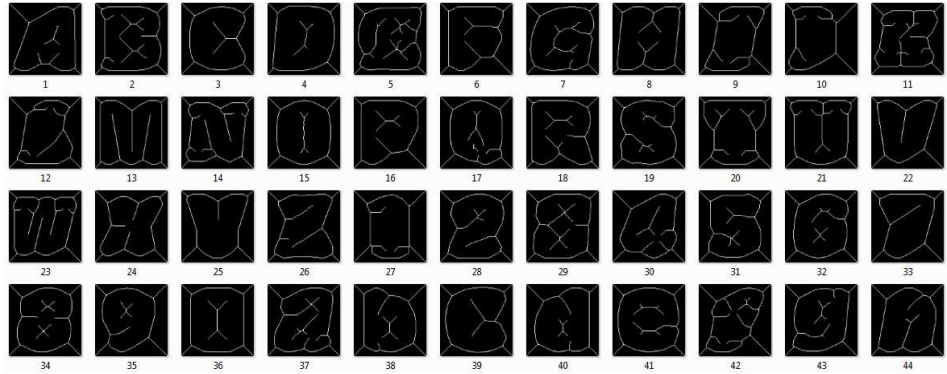}
\caption{The skeletonized database of subset of characters}
\label{fig3}
\end{figure}
\subsection{Template Matching}
Template matching is a natural approach to pattern classification\cite{Jain00statisticalpattern}. It involves
determining similarities between a given template and windows of the same size in an image and identifying the window that produces the highest similarity measure. It works by comparing derived image features of the image and the template for each possible displacement of the template. This process involves the use of a database of characters or templates. We have created a template for all possible input characters. The skeleton of all the characters in the dataset is refined to fit into a window without white spaces and the template is created. The templates are normalized to 42 x 24 pixels and stored in the database. The skeletonized database of such subset of characters are shown in Fig.~\ref{fig3}.
\subsection{Matching Strategy}
\begin{figure}[hbtp]
\centering
\includegraphics[scale=0.6]{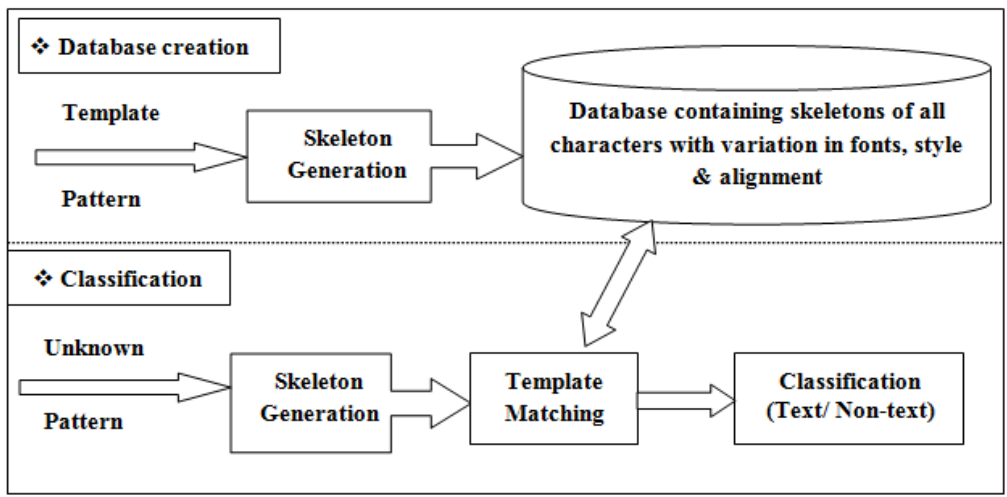}
\caption{Matching Strategy}
\label{fig4}
\end{figure}

\begin{figure}[hbtp]
\centering
\includegraphics[scale=0.6]{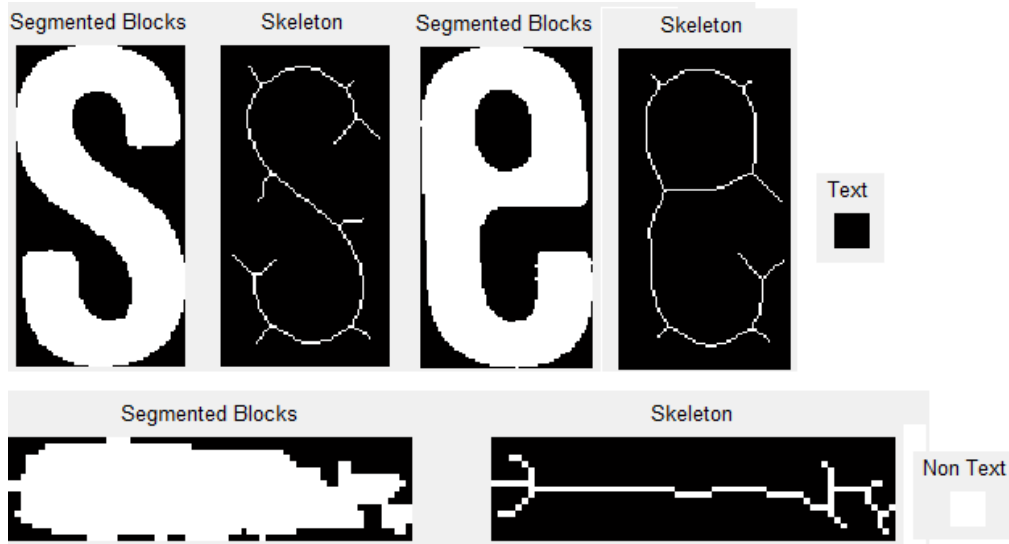}
\caption{Text classification of the segmented blocks using template matching}
\label{fig5}
\end{figure}

In the proposed method, text classification is done using template matching based on 2-D correlation coefficients between the characters as shown in Fig.~\ref{fig4}. 
During the process of classification, the current input character is compared to each template to find either an exact match or the template with the closest representation of the input character. If I$(x,y)$ is the input character, Tn$(x,y)$ is the template n, then the matching function s$(I,Tn)$ will return a value indicating how well the template $'$n$'$ matches the input character. Normalization is done through mask processing to view the transformation. This mapping is used to map every pixel of the original image to the corresponding pixel in the normalized image. After normalization, the skeletonized character of the input test image is further matched with all the skeletonized characters in the template database using 2-D normalized correlation coefficients approach to identify similar patterns between a test image and the standard database skeletonized images i.e.,
\begin{equation}
s\left(I,T_{n}\right) = \frac{\sum^{w}_{i=0}\sum^{h}_{j=0}\left( I\left( i,j\right) -\vert I \vert \right)\left( T _{n}\left( i,j\right) - \vert T_{n} \vert \right)}
{\sqrt{\sum_{i=0}^{w}\sum_{j=0}^{h}\left( I\left( i,j\right) -\vert I \vert \right)^2\left( T _{n}\left( i,j\right) - \vert T_{n} \vert \right)^2 }}
\end{equation}
This method is efficient and has high speed when dealing with character identification. For templates without strong features or when the bulk of the template image constitutes the matching image, a template-based approach may be effective. Since template-based matching may potentially require sampling of a large number of points, it is possible to reduce the number of sampling points by reducing the resolution of the search and template images by the same factor. The resized image of skeletonized characters (alphabets A-Z, a-z , numbers 0-9) is further used for formation of character templates. These character templates are in the form of feature vectors which are stored as reference data pattern. The reference data pattern is used at the time of template matching to the appropriate character. Once the template matching is performed, the template having the highest similarity is said to be highly correlated and hence it is considered as the best matched template. This template is further used for classifying whether the segmented block is a text block or a non-text block as illustrated in Fig.~\ref{fig5}.  
Character recognition is based on the previously constructed database which contains the important features related to the characters that are already known.  It shall be observed here that the database can be made self expandable to accommodate new font styles, new alphabet sets etc. to increase the localization accuracy. The skeleton of the characters in various font styles are stored in the knowledge base which is used for matching. In the classification phase, the learned prototypes are used to classify the unknown incoming patterns to the class of the matching prototype. In this phase, we extract features of the segmented block and compares these features with those recorded in the database. If the features are matched completely or closely matched, then the  segmented input block is classified into a known class (text blocks). Otherwise, the segmented block belongs to a class of non-text blocks as depicted in Fig.~\ref{fig5}.
\subsection{Text Localization}
The objective of text localization is to place rectangles of varying sizes covering the text regions\cite{YiFeng2009}. We start by merging all the text detections. We employ geometrical analysis to identify the text components and group them to localize text regions. The false positives are eliminated by computing height, width, aspect ratio$(AR)$ and using some geometrical rules devised based on edge area$(EA)$ of the text blocks.
 \begin{center}
          $ AR = width / height$  \\
              $ density = EA / ( height * width ) $
              \end{center}  
According to the attributes of the horizontal text line, we make the following rules to confirm on the non text blocks.
       \begin{center}
        $  i)  AR < T1  ||   density < T2  $ \\
        $  ii) height > 50 ||  height < 6  $  \\
        $ iii) width < 5 ||  height * width < 24 $ 
        \end{center}         
The candidate text lines are obtained by applying these rules. The thresholding values $T1$ and $T2$ are the calculated mean and standard deviation respectively. Then, we label the connected components by using 4-connectivity. The foreground connected components for each of these images are considered as text candidates. Further, the morphological dilation operation is performed to fill the gaps inside the obtained text regions which yields better results and the boundaries of text regions are identified.
\section{Experimental Results}
\begin{figure}
\includegraphics[scale=0.375]{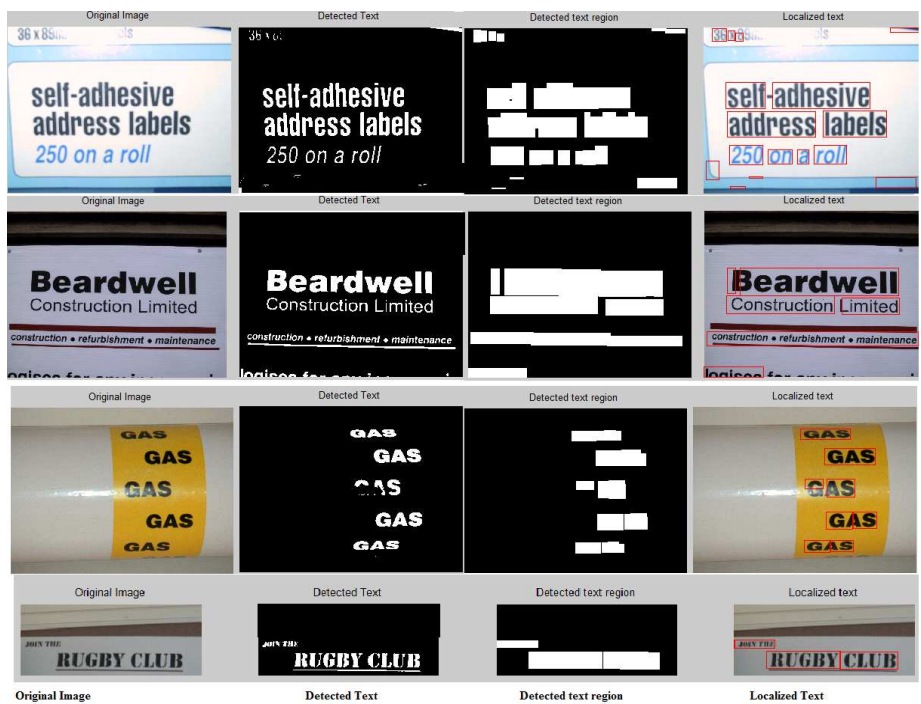}
\caption{Sample results of image localization}
\label{fig6}
\end{figure}

This section presents the experimental results to reveal the success of the proposed approach. The evaluation of the system on the ICDAR database shows that it is capable of detecting and locating texts of different sizes, styles and types present in natural scenes. The performance of the proposed approach is evaluated for scene images with respect to f-measure(F) which is a combination of two metrics:   precision(P) and recall(R) and   the results are reported in Table~\ref{tab1}. We have conducted experiments on ICDAR 2003 text locating competition dataset\cite{Lucas2005} and the localized text blocks  are shown in Fig.~\ref{fig6}. In order to exhibit the performance of the proposed approach, we have made a comparative study with the state of the art text localization approaches\cite{Pan2009}\cite{epshtein2010}.
\begin{table}
\centering
\caption{Evaluation performance for ICDAR 2003 dataset }
\begin{tabular}{|c|c|c|c|c|}
\hline 
 \cline{2-4}
Methods & R & P & F  \\ 
\hline 
 Pan et. al\cite{Pan2009} & 0.67 & 0.71 & 0.69  \\ 
\hline 
Epshtein et. al\cite{epshtein2010} & 0.73 & 0.60 & 0.66  \\ 
\hline 
Proposed & 0.83 & 0.79 &  0.81 \\
\hline
\end{tabular} 
\label{tab1}
\end{table}
\begin{table}
\centering
\caption{Evaluation performance for ICDAR 2011 dataset }
\begin{tabular}{|c|c|c|c|c|}
\hline 
 \cline{2-4}
Methods & R & P & F  \\ 
\hline 
Yi et. al\cite{Liu2012} & 0.71 & 0.67 & 0.62  \\ 
\hline 
Neumann et. al\cite{Neumann2012} & 0.65 & 0.63 & 0.69  \\ 
\hline
Proposed & 0.85 & 0.81 &  0.83 \\
\hline
\end{tabular} 
\label{tab2}
\end{table}
\\Pan et.al\cite{Pan2009} used a Conditional Random Field (CRF) model and energy minimization approach to detect and localize the text. However, this method was meaningful for unconstrained scene text localization. Epshtein et. al\cite{epshtein2010} employed the stroke width transform method, canny edge detector and connected component analysis for text localization. The limitation of the method is its dependency on successful edge detection which likely failed on blurred or low-contrast images.

The proposed method is also evaluated using the same parameters on ICDAR 2011 dataset and the results are reported in Table~\ref{tab2}. Yi et.al\cite{Liu2012} employed a text region detector, condition random field model and learning-based energy minimization approach to detect and localize the text in natural scene images. This approach obtained better recall rate but it was time consuming. Neumann et. al\cite{Neumann2012} achieved a good performance for character detection by using an efficient sequential selection method for characters from the set of Extremal Regions (ERs). The method fails against noise and low contrast of characters which was demonstrated by false positives that exist due to watermarked text.

\begin{figure}
\centering
\includegraphics[scale=0.4]{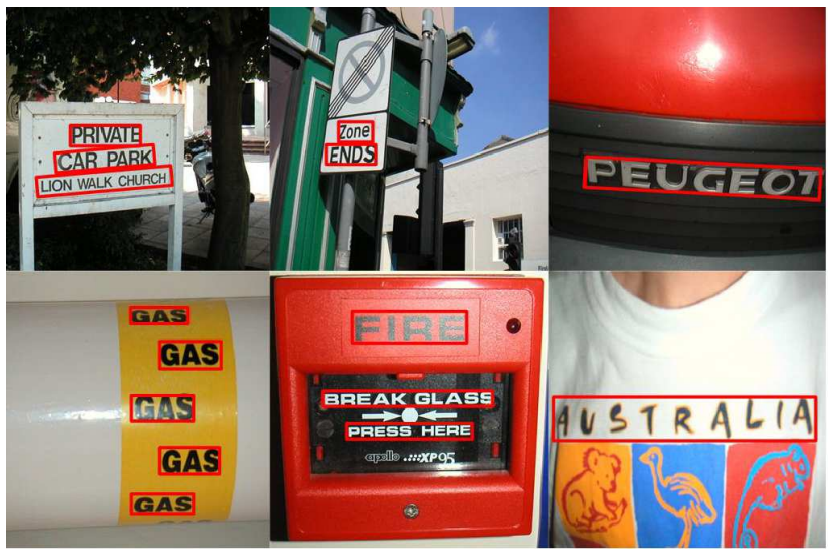}
\caption{Sample results of localization as found in literature\cite{Liu2012}}
\label{fig7}
\end{figure}

\begin{figure}
\centering
\includegraphics[scale=0.5]{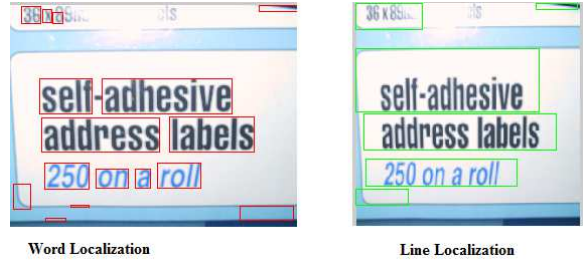}
\caption{Sample text localization results of the proposed approach if bounding boxes are placed word-wise and line-wise. }
\label{fig8}
\end{figure}

The proposed approach has gained slightly improved precision and recall rates on ICDAR 2011 dataset and comparable results for ICDAR 2003 dataset when compared to other approaches in the literature. For visualization purpose, we have shown the sample text localization results of others\cite{Liu2012} are shown in Fig.~\ref{fig7}. The sample text line localization results of our proposed method is shown in Fig.~\ref{fig8}. The evaluation performance for ICDAR dataset of the proposed method when text line is localized is highlighted in Table~\ref{tab3}.
\begin{table}
\centering
\caption{Evaluation performance of text line localization for the proposed method }
\begin{tabular}{|c|c|c|c|c|}
\hline 
 \cline{2-4}
Methods & R & P & F  \\ 
\hline 
ICDAR 2003 & 0.86 & 0.83 & 0.80  \\ 
\hline 
ICDAR 2011 & 0.84 & 0.79 & 0.82 \\ 
\hline
\end{tabular} 
\label{tab3}
\end{table}

\section{Conclusions}
\label{section:conclusions}

Text embedded in scene images contain abundant high level semantic information which is important to analysis, indexing and retrieval. We developed a skeleton matching based approach that classifies the text blocks through template matching. The newly developed approach is capable of localizing the text regions in  scene images.  Experimental results show that the proposed method is effective for identifying text and non text blocks.  The various problems that occur due to complex background need to be addressed in our future works.  

\section*{Acknowledgement}

Authors would like to thank Department of Science and Technology, Government of India for the financial support granted vide reference number INT/RFBR/P-133.

\end{document}